\documentclass[sigconf]{acmart}

\usepackage{booktabs} 
\usepackage{latexsym}
\usepackage{url}
\usepackage{graphicx}
\usepackage{mathtools}
\usepackage{multirow}
\usepackage{subcaption}
\usepackage{adjustbox}
\usepackage{enumitem}
\usepackage{array}
\usepackage{latexsym}
\usepackage{url}
\usepackage{graphicx}
\usepackage{mathtools}
\usepackage{multirow}
\usepackage{subcaption}
\usepackage{adjustbox}
\usepackage{enumitem}
\usepackage{array}
\usepackage{listings}

\usepackage[small]{titlesec}

\usepackage{listings}

\usepackage{color}
\definecolor{gray}{rgb}{0.4,0.4,0.4}
\definecolor{darkblue}{rgb}{0.0,0.0,0.6}
\definecolor{cyan}{rgb}{0.0,0.6,0.6}

\newcommand{\ignore}[1]{}
\newcommand{\squishlist}{
 \begin{list}{$\bullet$}
  { \setlength{\itemsep}{0pt}
     \setlength{\parsep}{2pt}
     \setlength{\topsep}{2pt}
     \setlength{\partopsep}{0pt}
     \setlength{\leftmargin}{1em}
     \setlength{\labelwidth}{1em}
     \setlength{\labelsep}{0.4em} } }

\newcommand{\squishend}{
  \end{list}  }

\lstdefinelanguage{XML}
{
  morestring=[b]",
  morestring=[s]{>}{<},
  morecomment=[s]{<?}{?>},
  stringstyle=\color{black},
  identifierstyle=\color{darkblue},
  keywordstyle=\color{cyan},
  morekeywords={VALID,ACCEPTABLE,INVALID,POSITIVE,NON_ENGAGING, ENGAGING, NEUTRAL}
}

\usepackage{xcolor}
\usepackage[normalem]{ulem}
\definecolor{darkgreen}{rgb}{0.0, 0.5, 0.0}

\def\OLdel#1{\bgroup\markoverwith{\textcolor{darkgreen}{\rule[0.5ex]{2pt}{1pt}}}\ULon{#1}}

\def\HH#1{{\color{cyan}HH: \it #1}}
\def\HHdel#1{\bgroup\markoverwith{\textcolor{blue}{\rule[0.5ex]{2pt}{1pt}}}\ULon{#1}}

\def\HH#1{\relax} 
\def\HHdel#1{#1} 

\definecolor{purple}{rgb}{0.5, 0.0, 0.5}

\def\ACdel#1{\bgroup\markoverwith{\textcolor{purple}{\rule[0.5ex]{2pt}{1pt}}}\ULon{#1}}

\setcopyright{rightsretained}
\acmPrice{15.00}
\acmDOI{10.1145/3139491.3139504}
\acmYear{2017}
\copyrightyear{2017}
\acmISBN{978-1-4503-5558-2/17/11}
\acmConference[ISIAA'17]{1st ACM SIGCHI International Workshop on Investigating Social Interactions with Artificial Agents}{November 13, 2017}{Glasgow, UK}

\begin{document}
\title{A Review of Evaluation Techniques for Social Dialogue Systems}

\subtitle{Extended Abstract}



\author{Amanda Cercas Curry}
\affiliation{%
\department{Interaction Lab}
  \institution{Heriot-Watt University}
  \city{Edinburgh} 
  \state{U.K.} 
  \postcode{EH14 4AS}
}
\email{ac293@hw.ac.uk}

\author{Helen Hastie}
\affiliation{%
\department{ Interaction Lab}
  \institution{Heriot-Watt University}
  \city{Edinburgh} 
  \state{U.K.} 
  \postcode{EH14 4AS}
}
\email{h.hastie@hw.ac.uk}

\author{Verena Rieser}
\affiliation{%
\department{Interaction Lab}
  \institution{Heriot-Watt University}
  \city{Edinburgh} 
  \state{U.K.} 
  \postcode{EH14 4AS}
}
\email{v.t.rieser@hw.ac.uk}


\renewcommand{\shortauthors}{A. Cercas Curry et al.}

\begin{abstract}

In contrast with goal-oriented dialogue, social dialogue has no clear measure of task success. Consequently, evaluation of these systems is notoriously hard. In this paper, we review current evaluation methods, focusing on automatic metrics.  We conclude that turn-based metrics often ignore the context and do not account for the fact that several replies are valid, while end-of-dialogue rewards are mainly hand-crafted. Both lack grounding in human perceptions. 

\end{abstract}

%
%

\ccsdesc[500]{Human-centered computing~Natural language interfaces}
\ccsdesc[300]{General and reference~Surveys and overviews}
\ccsdesc[300]{General and reference~Evaluation}

\keywords{Automatic Evaluation, Social Dialogue Systems, Conversational Agents, Evaluation Metrics}

\maketitle

\section{Introduction}

Non-task-oriented, social dialogue systems, aka ``chatbots", receive an increasing amount of attention 
 as they are designed to establish a \textit{rapport} with the user or customer, providing engaging and coherent  dialogue. Traditional dialogue systems \cite{mctear:2004, rl11book} tend to be task-orientated for a limited domain and evaluation  methods of such systems have been much researched (see \cite{hastie2012} for an overview). Evaluation of social dialogue systems, on the other hand, is challenging as there is no clear measure for task success and evaluating whether such a \textit{rapport} has been established is far from clear-cut.  
\ignore{
Methods for developing social dialogue systems include:
neural approaches, e.g.\ \cite{vinyals2015neural,sordoni-EtAl:2015:NAACL-HLT}; 
Machine Translation (MT),  e.g.\ \cite{ritter:2011};  Information Retrieval, e.g.\ \cite{Banchs:acl2012}; deep Reinforcement Learning approaches, e.g.\ \cite{Li2016}; and  hybrid models incorporating handwritten rules, e.g. \cite{ticktock:sigdial16}.
}
One common method for evaluating such systems is human evaluation where subjects are recruited to interact with and rate different systems. However, human evaluation is highly subjective, time-consuming, expensive and requires careful design of the experimental set-up. 

\ignore{
\section{Human Evaluation}\label{ssec:human}
The following metrics are obtained through hand-annotation either by experts/wizards or provided by participants, either during or after interacting with the system\footnote{Please note that this list is non-exhaustive, due to the sheer variety of different human evaluation performed.}:
\squishlist
\item Ratings or Likert scale judgements of e.g.\ Naturalness \cite{Shang2015}, Fluency \cite{Serban2016}, 
Adequacy/Clarity of reference \cite{Li2016}, Informativeness \cite{Wen2015} either at system or turn level,
\item Coherence, through annotations of turn-level appropriateness, measuring the validity of a response in a given context
e.g. {\em valid}, {\em acceptable}, {\em invalid} \cite{wochat,Yu2016}, 
\item Engagement, as annotated from video at the turn level and overall dialogue engagement (both self-assessed and by experts) \cite{yuthesis}. 
\item Pairwise response choice, i.e. choose one response over another \cite{sordoni-EtAl:2015:NAACL-HLT},

\squishend

As mentioned above, evaluation using human judges can be costly and requires careful engineering of the task instructions in order to control bias. Data collection analysis can be slow and hinder the development process. 
The rise of crowdsourcing has lowered these data collection costs, however, it is not without its own downfalls \cite{crowdsourcing:2013}. For instance, it is hard to ensure the legitimacy of the ratings as crowdworkers may not be native speakers or may try to get through a job as fast a possible without paying adequate attention to the job instructions or quality of responses. Whilst measures such as limiting the geographic location of the crowdworkers or adding ``test questions" can help prevent this, it remains a challenge to collect ``clean" data through crowdsourcing.
Note that for task-based systems there are attempts to evaluate systems ``in the wild", i.e. testing with real customers who have real goals (as opposed to paid subjects), e.g.\ \cite{raux2005let,gruenstein:2007}.
For social interaction, the closest equivalent to this is the the Amazon Alexa Challenge\footnote{\url{https://developer.amazon.com/alexaprize}}, where Amazon customers get to rate the overall experience with the system. The challenge aim is to  produce conversations that are engaging enough for the user to chat for at least 20 minutes. This contrasts with the Turing Test and Loebner Prize\footnote{\url{http://www.loebner.net/Prizef/loebner-prize.html}}, where the goal is for the system to behave indistinguishable from a human.

}

\section{Automatic Metrics}
\ignore{
Automatic evaluation has been previously studied for traditional spoken dialogue systems and has been used for optimisation functions in adaptive systems \cite{rieser2008automatic}.
These methods are based on, for example, on the PARADISE method \cite{paradise} e.g.  \cite{hastie2002trouble} or the notion of a Mental model as in the MeMo system \cite{Moller_memo} or through statistical models such as HMMs \cite{engelbrecht}.
}
Automatic evaluation is popular because it is cost-effective and faster to run than human evaluation, and is needed for automatic benchmarking and tuning of algorithms. Here, we  discuss  existing automatic methods for developing social systems in terms of word-overlap metrics, machine learning-based estimation models and reward-based metrics. 
Since social systems lack a final success measure, many of the discussed metrics operate at turn-level.

\subsection{Word-Overlap Metrics}
Word-overlap metrics, such as BLEU \cite{BLEU} and ROUGE \cite{lin2004rouge}, are borrowed from Machine Translation (MT) and Summarisation 
and have been widely been used to evaluate neural dialogue system output, as reported in, for example \cite{Li2016,sordoni-EtAl:2015:NAACL-HLT}. 
However, these metrics have not been shown to correlate well with human judgements in a dialogue setting \cite{Liu:EMNLP2016}. One possible explanation is that there is no  ``gold standard" to compare with, as in MT: 
there may be many valid responses to an utterance that have no or few overlapping n-grams and would thus receive low BLEU or ROUGE scores, see example in Table \ref{tab:example}.    Measures from information theory such as perplexity have also been used for evaluation, e.g. comparing neural models to n-grams \cite{vinyals2015neural}, however perplexity can be difficult to interpret. There is, therefore, a need for an evaluation method that does not measure success by comparing an utterance to human-generated responses but instead considers the utterance itself and its appropriateness within its context. 

\begin{table}
\caption{Valid system response with low word overlap to reference.
\vspace{-.25cm}}\label{tab:example}
\begin{tabular}{m{18em}} \label{bleuexample}
\\\hline
  User utterance:\\ 
  \textit{Have you read Murakami's new novel?}\\\hline
  Reference response:\\ 
  \textit{No I don't think I have read Murakami's new novel, what is it about?}\\\hline
  System output: \\ 
  \textit{Yes, it wasn't my favourite but I still liked it.}\\\hline
\end{tabular}
\vspace{-.25cm}
\end{table}

\subsection{Machine Learning Methods for Dialogue Evaluation}
Recently, Machine Learning (ML) based evaluation has gained popularity. This method operates on the turn-level and aims to provide an estimation model of a ``good" response. The advantages of this method is that it has been shown to come closer to human-generated responses \cite{lowe:2017} than BLEU and ROUGE. However, 
 such methods require retraining for each domain.
\paragraph{Discriminative Models:}
These models attempt to distinguish the ``right" from the ``wrong" answer.
Next-Utterance Classification (NUC) \cite{Lowe:NUC16} can be evaluated by measuring the system's ability to select the next answer from a list of possible answers sampled from elsewhere in the corpus, using retrieval metrics such as recall.
NUC offers several advantages: performance is easy to compute automatically and the task is interpretable and can be easily compared to human performance. 
However, 
similar issues to word-based metrics do apply in that there is not necessarily one single correct answer. 

More recently, adversarial evaluation measures have been proposed to distinguish a dialogue model's output from that of a human. For example, the model proposed by \cite{DBLP:journals/corr/KannanV17} achieves a 62.5\% success rate using a Recurrent Neural Networks (RNN) trained on email replies. 

\paragraph{Classification Models:}
\cite{lowe:2017} propose to predict human scores from a large dataset of human ratings of Twitter responses. The proposed models learn distributed representations of the context, reference response and the system's response using a hierarchical RNN encoder. 
The learned model correlates with human scores at the turn level and also generalises to unseen data. However, it does tend to have a bias towards generic responses.

\subsection{Reward-based Metrics}

Reinforcement Learning (RL) based models have been applied to task-based systems \cite{rl11book} to optimise interaction for some \textit{reward}. For social systems, this has also been investigated as a means to avoid generic responses, such as ``I don't know". Here, the evaluation function is implemented as the reward. We will discuss these types of reward at  turn-level and at  system-level.  

\paragraph{Turn-level rewards:} \cite{Li2016} propose a metric involving a weighted sum of three measures:
\squishlist
\item Coherence: semantic similarity between consecutive turns,
\item Information flow: semantic dissimilarity between utterances of the same speaker,
\item Ease of answering: negative log-likelihood of responding to an utterance with a dull response (as defined by a blacklist).
\squishend
 In their experiments, they find the RL approach outperforms their other systems in terms of dialogue length, diversity of answers and overall quality of multi-turn dialogues. This suggests that the proposed reward function successfully captures the relationship between an utterance and a response at least partially, which can be useful in evaluating potential responses without the need for human-generated references. However,  while coherence at the turn-level is a key factor in quality estimation, it does not necessarily reflect the overall quality of the dialogue. 

\paragraph{System-level rewards:}
The reward function by \cite{Li2016} was based on heuristics, whereas \cite{ticktock:sigdial16} use a Wizard-of-Oz experiment to measure engagement and deduct a reward function with the following metrics:
\squishlist
\item Conversational depth: the number of consecutive turns belonging to the same topic, 
\item Lexical diversity/information gain: the number of unique words that are introduced into the conversation from both the system and the user,
\item Overall dialogue length.
\squishend


\ignore{

\section{Inter-disciplinary Notions of Engagement}

The levels of engagement as a turn-level measure can be used to have the system adapt, e.g. by changing the topic \cite{Yu2016Woz}. However, overall engagement in these chatbot systems, including future-use, is also an important measure.  An interesting use-case is Microsoft's Chinese Xiaoice chatbot \ that has over 20 million registered users with the average user interacting with the service 60 times a month \cite{xiaoice}.


\cite{sidner03} define engagement as ``the process by which two (or more) participants establish, maintain and end their perceived connection. This process includes: initial contact, negotiating a collaboration, checking that other is still taking part in the interaction, evaluating whether to stay involved, and deciding when to end the connection". 

Engagement, immersion and intention are clearly important aspects for social dialogue systems in terms of enjoyment and user satisfaction but also in terms of continued and repeated use, particularly so for the large number of emerging commercial systems. Measuring engagement, however,  is particularly challenging for dialogue systems, more so than in the field of human-robot interaction \cite{Castellano13}, intelligent tutoring systems \cite{litman:itspoke}, gaming \cite{jennett08} and multimodal interfaces. For example \cite{yuthesis} shows a significant improvement in engagement detection when visual cues are included as features in a multimodal interface with an F1-score of 0.8 for a binary engaged/not engaged. 
In the gaming context, engagement is measured in terms of the number of clicks, mouse movements, time spent looking at the screen, etc. 
These types of input features, however, are not available in the context of speech-only and text-based spoken dialogue systems.  We, therefore, need approximations of user engagement that rely solely on features accessible though speech/text such as: the number of turns, user reactivity, sentiment polarity, self-reported engagement as well as taking advantage of the large body of work on emotion recognition to approximate engagement e.g. as in \cite{Abadi13}.


Linked to immersion and engagement is the concept of \textit{Flow}, defined by 
Csijszentmihalyi  as ``the mental state of operation in which a person performing an activity is fully immersed in a feeling of energized focus, full involvement, and enjoyment in the process of the activity". To achieve this sense of flow, one must be involved in an activity with a clear set of goals and progress, thus, adding direction and structure to the task. The user should also have clear and immediate feedback, thus allowing  them to adjust their performance to maintain the flow state.
Finally, one must have confidence in one's ability to complete the task at hand \cite{flow}. As such, social and task-based systems should support the user in his/her goal, give frequent feedback/progress and increase user confidence. Automatically measuring a system's effectiveness to maintain flow is challenging given these multiple dimensions.

}
\section{Conclusion and Discussion}
It is clear that there is still work to be done with respect to establishing an effective evaluation method that can capture all aspects of dialogue from naturalness and coherence to long-term engagement and flow. Word-based metrics such as BLEU, ignore the 
fact there may be any number of equally valid and appropriate responses, 
while turn-based metrics 
 do not account for the over-use of generic responses, 
and system-level rewards are based on heuristics. 
In future work, we will utilise data we gathered as part of the Amazon Alexa Prize challenge to build a data-driven model to predict customer ratings.   



\footnotesize
\section*{Acknowledgements}
This research is supported by Rieser's EPSRC projects DILiGENt (EP/M005429/1) and   MaDrIgAL (EP/N017536/1); and from the RAEng/Leverhulme Trust Senior Research Fellowship Scheme (Hastie/LTSRF1617/13/37).


\bibliographystyle{ACM-Reference-Format}
\bibliography{acl2017} 

\end{document}